%% file: main.tex
\documentclass{article}
\usepackage{arxiv}

\usepackage[utf8]{inputenc} 
\usepackage[T1]{fontenc}    

\usepackage{graphicx}%
\usepackage{multirow}%
\usepackage{amsmath,amssymb,amsfonts}%
\usepackage{amsthm}%
\usepackage{mathrsfs}%
\usepackage[title]{appendix}%
\usepackage{xcolor}%
\usepackage{textcomp}%
\usepackage{manyfoot}%
\usepackage{booktabs}%
\usepackage{algorithm}%
\usepackage{algorithmicx}%
\usepackage{algpseudocode}%
\usepackage{listings}%
\usepackage{xurl}
\usepackage[nameinlink,noabbrev]{cleveref}
\usepackage{isomath}
\usepackage{tikz}
\usepackage{csquotes}
\usepackage{multirow}

\title{Traversing the Subspace of Adversarial Patches}

\author{
    Jens Bayer \\
        Fraunhofer Center for Machine Learning,\\
        Fraunhofer IOSB\\
        Gutleuthausstr. 1, Ettlingen, Germany\\
        \texttt{jens.bayer@iosb.fraunhofer.de} \\
    \And
        Stefan Becker \\
        Fraunhofer IOSB\\
        Gutleuthausstr. 1, Ettlingen, Germany\\
    \And
        David M\"unch \\
        Fraunhofer IOSB\\
        Gutleuthausstr. 1, Ettlingen, Germany\\
    \And
        Michael Arens \\
        Fraunhofer IOSB\\
        Gutleuthausstr. 1, Ettlingen, Germany\\
    \And
        J\"urgen Beyerer \\
        Karlsruher Institute of Technology,\\
        Fraunhofer Center for Machine Learning,\\
        Fraunhofer IOSB\\
        Fraunhoferstr. 1, Karlsruhe, Germany\\
}

\begin{document}
\maketitle


\begin{abstract}Despite ongoing research on the topic of adversarial examples in deep learning for computer vision,
some fundamentals of the nature of these attacks remain unclear. As the manifold hypothesis posits,
high-dimensional data tends to be part of a low-dimensional manifold. To verify the thesis with
adversarial patches, this paper provides an analysis of a set of adversarial patches and
investigates  the reconstruction abilities of three different dimensionality reduction methods.
Quantitatively, the performance of reconstructed patches in an attack setting is measured and the
impact of sampled patches from the latent space during adversarial training is investigated. The
evaluation is performed on two publicly available datasets for person detection. The results
indicate that more sophisticated dimensionality reduction methods offer no advantages over a simple
principal component analysis.
\end{abstract}



\input{src/introduction}
\input{src/relatedwork}
\input{src/recap}
\input{src/experiments}

\input{src/conclusion}
\begin{appendices}



\newpage
\section{Additional Data}
\input{src/appendix}

\end{appendices}

\section*{Acknowledgements}
This work was developed in Fraunhofer Cluster of Excellence \enquote{Cognitive Internet Technologies}.
\newpage

\bibliographystyle{unsrt}
\bibliography{src/bibliography}

\end{document}

%% file: src/introduction.tex
\section{Introduction}
Adversarial patch attacks in deep learning for computer vision are a quite well-known topic, yet,
some fundamentals of the nature of these powerful attacks remain unclear. Due to the complexity of
deep neural networks, where these patches are optimized on, there are no straightforward
explanations why a generated pattern, looks the way it does.  Moreover, there are no explainability
methods like those used for deep neural networks that provide a human-understandable explanation
of why these patches prevent a network from working properly. To further understand these
high-dimensional data, this paper follows the manifold hypothesis~\cite{Fefferman2016}. For
adversarial attacks, this states, that adversarial patterns are part of a lower-dimensional
manifold. This paper builds on our recently published work \enquote{Eigenpatches -- Adversarial
Patches from Principal Components}~\cite{Bayer2023}. By applying a principal component analysis
(PCA) on a set of the trained adversarial patches, it could already be shown that linear
combinations of Eigenpatches can be used to successfully attack the investigated YOLOv7 object
detector~\cite{Wang2022}.
\newpage

The contributions of this paper are:
\begin{enumerate}
\item[(i)] A more in-depth analysis of a crafted set of adversarial patches. 
\item[(ii)] An evaluation of the performance of patches sampled from different low-dimensional
	manifolds and further analyze their generalization ability by evaluating them across varying
	detection models and datasets.
\item[(iii)] An evaluation of the impact of the usage of sampled adversarial patches from those
    manifolds when used in adversarial training.
\end{enumerate}

The structure of the paper is as follows: In \Cref{sec:relatedwork} the similarities and
distinctions between related work and this paper are presented. \Cref{sec:recap} covers the
fundamentals of eigenpatches and presents the alternative manifold learning methods that are
evaluated. The experiments and the results are described in \Cref{sec:setup}. A discussion,
followed by a brief conclusion and outlook, is given in \Cref{sec:conclusion}.

%% file: src/relatedwork.tex
\section{Related work}\label{sec:relatedwork}
\label{sec:related_work}
Despite being an active research area, the focus of analyzing adversarial patterns is
rarely~\cite{Tarchoun2023} on object detector methods and adversarial patches. Instead, the
investigations are performed mostly on image classifiers and attacks that induce high-frequent noise
across the whole
image~\cite{Tramer2017,Wang2019,Shi2021,Dohmatob2022,Shafahi2019a,Weng2023,Garcia2023}. Therefore
only some selected works on analyzing adversarial attack patterns and sampling from low-dimensional
embeddings are presented. For an overview of the different approaches to adversarial attacks, we
refer to these surveys~\cite{Akhtar2018,Chakraborty2021,Pauling2022,Wang2023}.

Wang et al.~\cite{Wang2019} propose a fast black-box adversarial attack that identifies key
differences between different classes using a PCA. The principal components are then used to
manipulate a sample into either a target class or the nearest other class.

Similarly, yet different, \emph{Energy Attack} from Shi et al.~\cite{Shi2021} leverages PCA to
obtain the energy distribution of perturbations generated by white-box attacks on a surrogate model.
This transfer-based black-box adversarial attack samples patches according to the energy
distribution, tiles them, and applies them to the target image. The extracted patches are
high-frequent noise and are used to attack image classifiers. Despite the name, they should
therefore not be confused with adversarial patches that are used to attack object detectors in a
physical world attack.

Another approach uses the singular value decomposition and is presented by Weng et
al.~\cite{Weng2023}. The authors combine the top-1 decomposed singular value-associated features for 
computing the output logits with the original logits, used to optimize adversarial examples. This
results in an improvement in the transferability of the attacks.

Regarding the subspace of adversarial examples, researches have explored methods to estimate the
dimensionality of the space of adversarial inputs. 

Dohmatob et al.~\cite{Dohmatob2022}, for example, investigate the vulnerability of neural networks
to black-box attacks, specifically examining low-dimensional adversarial perturbations. They found
that adversarial perturbations are likely to exist in low-dimensional subspaces that are much
smaller than the dimension of the image space, supporting the manifold hypothesis.

An explicit estimation of the dimensionality of shared adversarial subspaces of, e.g., two fully
connected networks trained on two different datasets, is presented in~\cite{Tramer2017}. By
examining untargeted misclassification attacks they demonstrate that manipulating a data point to
cross a model's decision boundary is likely to result in similar performance degradation when
applied to other models.

Tarchoun et al.~\cite{Tarchoun2023} recently studied adversarial patches from an
information theory perspective, measuring the entropy of random crops of the patches. Their findings
indicate that the mean entropy of adversarial patches is higher than in natural images. Based on
these results, they developed a defense mechanism against adversarial patches.

Moreover, theoretical limits on the susceptibility of classifiers to adversarial attacks are
demonstrated by Shafahi et al.~\cite{Shafahi2019a} using a unit sphere and unit cube.  They suggest
that these bounds may be potentially bypassed by employing extremely large values for the class
densitiy functions. Furthermore, their findings suggest that the fundamental limits of adversarial
training for specific datasets with complex image classes in high-dimensional spaces are far worse
than one expects.

Godfrey et al.~\cite{Godfrey2023} conducted further research on the relationship between adversarial
vulnerability and the number of perturbed dimensions. Their findings support the hypothesis that
adversarial examples are a result of the locally linear behavior of neural networks with
high-dimensional input spaces.

While the related works mainly focus on adversarial examples for image classifiers, this work
explores commonalities of adversarial patch attacks against object detectors. The investigated
dimensionality reduction methods are evaluated in an attack setting and also tested if sampled
patches can be used in adversarial training.

%% file: src/recap.tex
\section{Fundamentals}\label{sec:recap}
The following section covers the necessary theoretical backgrounds of the investigated
dimensionality reduction methods and manifold learning techniques. To be more precise, Eigenpatches
and autoencoders are investigated. Both methods offer an embedding of the data in a low-dimensional
space while also providing a simple sampling strategy.

\subsection{Eigenpatches}
Eigenpatches or Eigenimages~\cite{Sirovich1987} are calculated on a set of adversarial
patches~\cite{Bayer2023}. The term Eigenimages is the name of the eigenvectors that can be derived
from a set of training images when a PCA is applied. In general, Eigenimages can be used to
represent the original training images and recreate these through a linear combination of the
low-dimensional representation.

Given a set of adversarial patches 
\begin{equation}
    \mathcal{P} = \{ \tensorsym{P}_i | i = 1, \ldots, n \}, \quad \tensorsym{P}_i \in \left[0,1\right]^{C \times H \times W}
\end{equation}
where $H$ is the height in pixel, $W$ is the width in pixel and $C$ is the number of channels. A principal component analysis
is performed on $\mathcal{P}$. With the top $k$ principal components $\tensorsym{E}_j, j \in \{1, \ldots, k\}$ and the weights
$\lambda_{i,j}$, the set 
\begin{equation}
    \mathcal{\hat{P}} = \{ \tensorsym{\hat{P}}_i | i = 1, \ldots, n\} \quad \tensorsym{\hat{P}}_i = \sum_{j=1}^k \lambda_{i,j} \tensorsym{E}_j
\end{equation}
can be generated that consists of linear combinations of the principal components and is a
recreation of $\mathcal{P}$~\cite{Bayer2023}.

\begin{figure}
    \centering
    \includegraphics[width=0.6\textwidth]{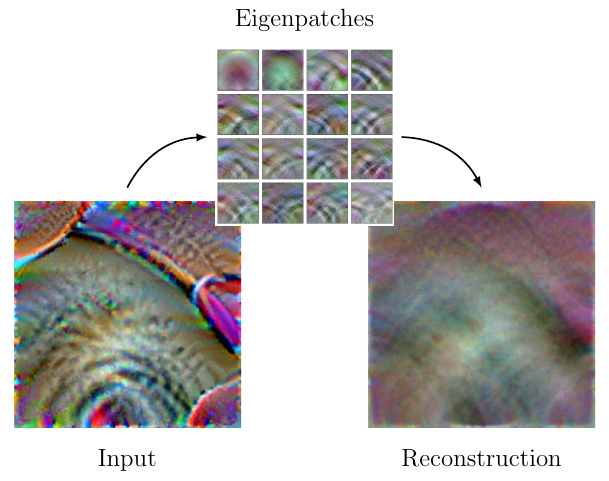}
    \caption{Reconstruction (right) of an adversarial patch (left) using the first 16 Eigenpatches
    (middle)~\cite{Bayer2023}.}
    \label{Eigenpatches}
\end{figure}

In some cases, especially when the data lies on or near a low-dimensional manifold within the
high-dimensional space, a PCA can be considered a way to approximate that manifold. However, a PCA
does not explicitly model the manifold structure of the data. It operates under the assumption, that
the principal components capture the most important directions in the data, but it doesn't take into
account the non-linear relationships or intrinsic geometry that may exist in the data.

To also take non-linearities into account, alternative methods are required. One particular
set of manifold learning techniques that are capable of extracting underlying structures from
high-dimensional data and projecting these onto a lower-dimensional space are autoencoders.

\subsection{Autoencoders}
Autoencoders are particularly valuable in manifold learning due to their ability to learn compact
and meaningful representations of high-dimensional data~\cite{Berahmand2024}. By compressing the
data into a lower dimensional latent space, autoencoders effectively capture the essential features
and structure of the data manifold. In general, autoencoders first encode the input data and decode
them after they were propagated through a bottleneck. During training, the reconstruction loss is
utilized to train the network weights. To capture spatial relation within image data, convolutional
autoencoders are used, as they replace the fully connected layers with convolutional layers in both
the encoder and decoder~\cite{Berahmand2024}.

An extension that introduces a probabilistic approach on learning the latent space representation
are variational autoencoders (VAE). They model the latent space as a probability distribution,
allowing for more flexible generation of new data points~\cite{Sohn2015}. In addition to the
reconstruction loss, the Kullback-Leibler divergence is calculated and used as a regularization term
to encourage the model to not focus on perfect reconstructions but rather be good at creating new
data~\cite{Kingma2014}.

The experiments described in the next section use a convolutional autoencoder and a conditional
variational autoencoder to generate reconstructions of the training data and sample new adversarial
patches. The conditional variational autoencoder is a variant of the variational autoencoder that
uses additional information during the encoding and decoding to condition the probability
distribution~\cite{Berahmand2024}.

%% file: src/experiments.tex
\section{Experiments}\label{sec:setup}
\subsection{Setup}
\begin{figure}
    \centering
    \includegraphics[height=0.9\textheight]{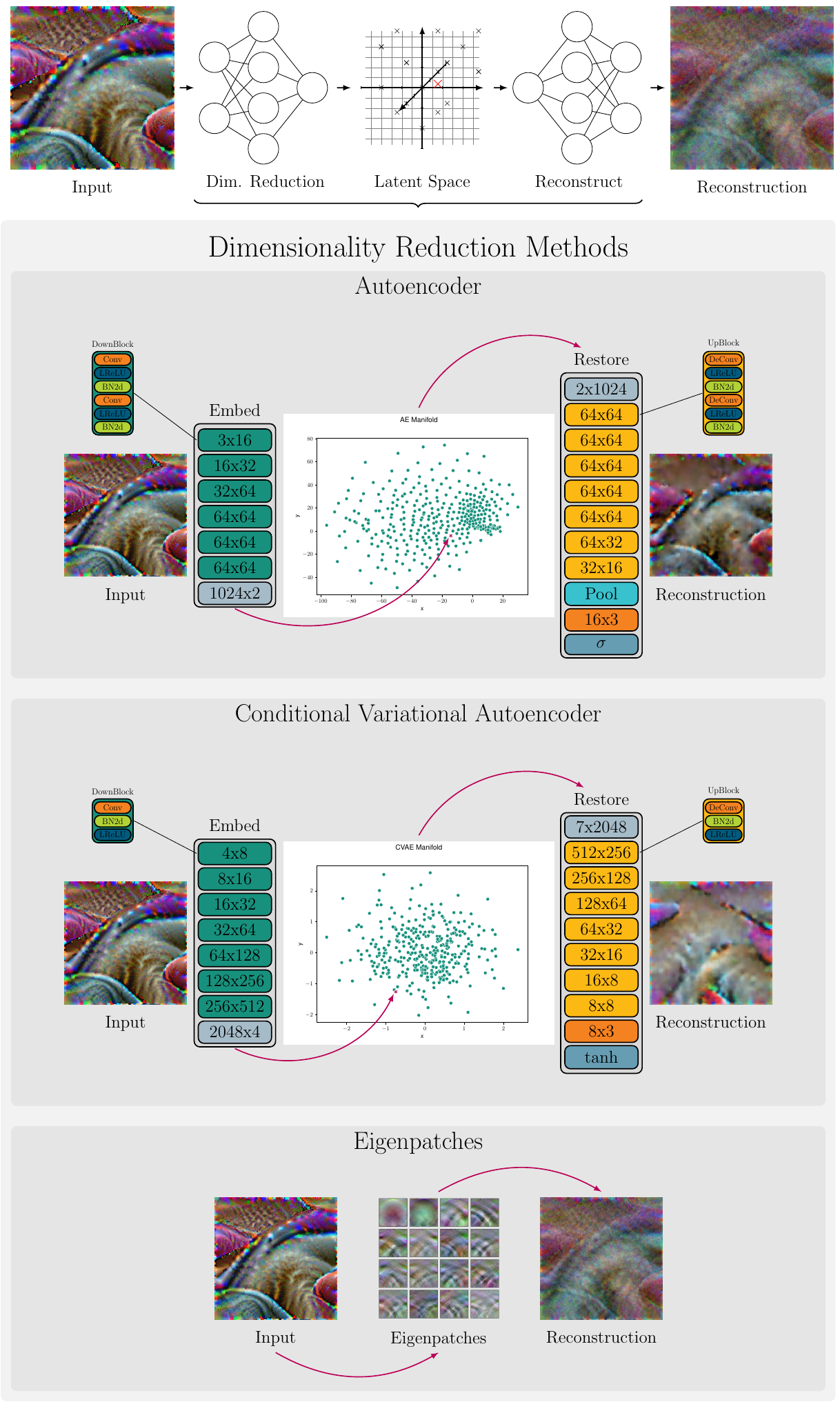}
    \caption{A schematic overview of the three dimensionality reduction methods. In general, the
    input is a prime patch, which will be encoded by a dimensionality reduction method. Both
    autoencoder approaches have a latent space with two dimensions. The number of dimensions for the
    PCA depends on the chosen number of principal components and varies between 16, 32, 64, and 128
    dimensions.  Based on the encoding in the latent space, the input is reconstructed.}
    \label{fig:experimental_setup}
\end{figure}

In the following, the experiments and the three different dimensionality reduction methods (see
\Cref{fig:experimental_setup}) are described and evaluated.

If not stated otherwise, all experiments use the YOLOv7 tiny model as architecture. Furthermore,
this paper uses the same patch set as \cite{Bayer2023}. They are referred as \emph{prime patches} in
the following. The prime patches are trained with different combinations (A-E) of rotation, scale
and lr-scheduler parameters as described in \cite{Bayer2023} (see \Cref{tbl:parameters}). They are
optimized on the YOLOv7 tiny model with the provided pretrained weights and the \emph{INRIA Person}
dataset.

To measure the impact on the performance of the detector, the mean average precision (mAP) is used.
The up/down facing arrows ($\uparrow$/$\downarrow$) in the tables indicate that a higher/lower
score is more desirable. This is particularly important to keep in mind, since a high mAP is desired
in the case of adversarial training and a low mAP in the case of a successful reconstruction of a
prime patch.

\subsubsection{Object Detector}
As in \cite{Bayer2023}, YOLOv7 is used as a reference object detector. For the evaluation of
adversarial training with patch reconstructions, the same training procedure is used for each
trained model. SGD with the default parameters to train the model from scratch provided by the
official git repository\footnote{\url{https://github.com/WongKinYiu/yolov7}} is used. The batch
size is set to 32 and the probability, for a bounding box to contain a patch, is set to $\pi=0.25$.

The single-patched network is trained with only a single prime patch and has not seen other patches
during training. The multi-patched networks are trained with multiple patches and have therefore
seen various patches during training. If an image is patched in a multi-patched network, all
boxes in the image share the same patch.

The prime multi-patched network is trained with 10 randomly predefined prime patches (see
\Cref{fig:predefined_prime_patches}). 

The PCA multi-patched networks are trained with linear combinations of Eigenpatches. The weights for
the linear combination are sampled from normal distributions with means and standard deviations
calculated regarding the encoded prime patches. 

Patches used in the convolutional autoencoder multi-patched network training are sampled by
uniformly selecting a random point in the latent space, bounded by the values of the encoded prime
patches. 

Similarly, the patches used in the training of the variational autoencoder multi-patched network 
are sampled. Here, values in the latent space are sampled from the underlying normal distribution,
conditioned to a randomly chosen parameter group.

\subsubsection{Datasets}\label{sec:datasets}
For the experiments, the \emph{INRIA Person} dataset~\cite{Dalal2005} and the
\emph{Crowdhuman} dataset~\cite{Shao2018} are used. Both datasets contain images of persons in
various environments.

The \emph{INRIA Person} dataset is a small dataset where the selected subset contains a total of
$585$ train images with $3\,317$ bounding boxes and $288$ test images with $904$ bounding boxes.
The dataset has also been used to generate the prime patches.

To verify the experiments, the \emph{Crowdhuman} dataset is used, which is a benchmark dataset for
person detection. The dataset contains $15\,000$ images in the training set, $4\,370$ in the
validation set and $5\,000$ images in the test set. The total number of human instances in all sets
is $470\,000$. For our experiments, we use the train set and validation set. Since there are
numerous small bounding boxes, each bounding box with less than $4\,096$ pixels is filtered and
removed from the dataset. This reduces the number of bounding boxes in the training set from
$332\,914$ to $109\,836$ and in the validation set from $97\,529$ to $32\,392$.
    
\subsubsection{Autoencoders}
A schematic of the architecture of both autoencoder models can be found in
\Cref{fig:experimental_setup}. Both models are trained over $2\,000$ epochs on the prime patches
with an initial learning rate of $0.01$ and the \emph{AdamW}~\cite{Loshchilov2019} optimizer. Each 100
epochs, the learning rate is reduced by a factor of 10. The batch size for both models is set to 64
and the bottleneck size is set to 2. For both models, the mean squared error between the input and
output is optimized. The variational autoencoder also optimizes the KL-Div loss.

\input{src/exp1}
\input{src/exp2}
\input{src/exp3}

%% file: src/exp1.tex
\subsection{T-Distributed Stochastic Neighbor Embeddings}
\begin{figure}[tb]
\centering
\includegraphics[width=\textwidth]{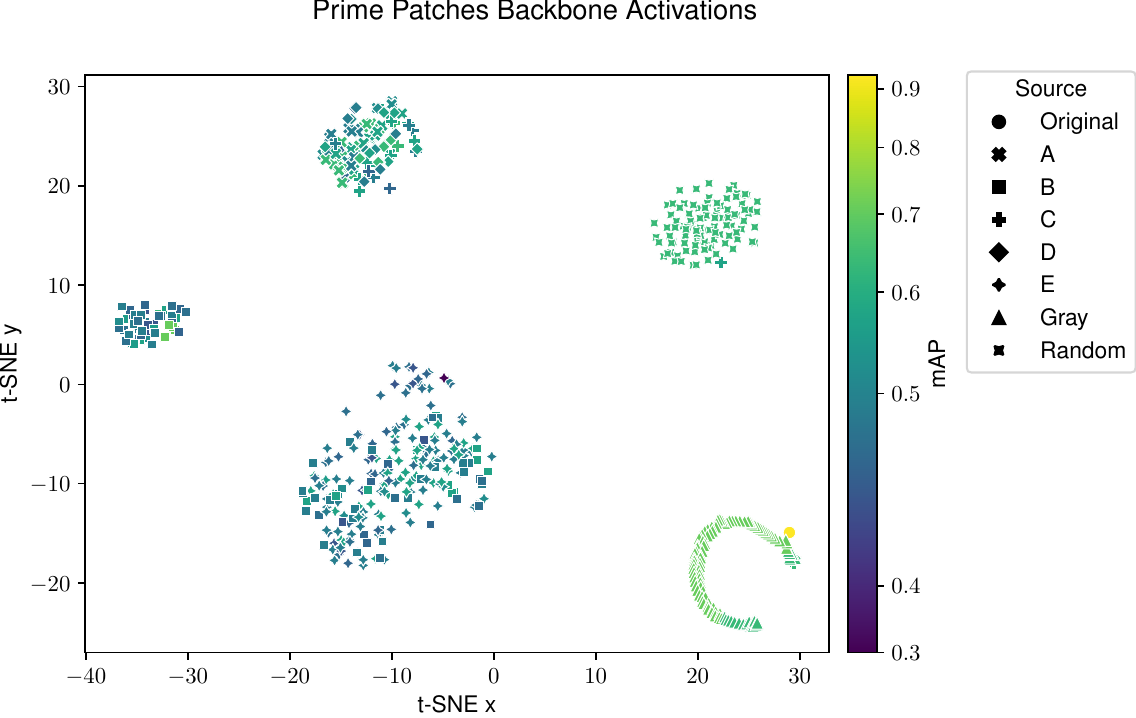}
\caption{t-SNE plot of the model backbone activation when attacked with a certain patch. 
Each marker represents a single patch, embedded at the same position in the same image. The color 
indicates the impact of the patch on the mean average precision. The parameter set of the patch is 
given by its marker symbol. The plot includes the prime patches, 100 random sampled patches as well
as 100 grayscale patches. The parameter settings (A-E) used to train a certain prime patch
correspond to the parameters presented in~\cite{Bayer2023}.}\label{fig:patch_activation_embedding}
\end{figure}

The prime patches and their influence on the activation of the last convolutional layer of the 
backbone layer in the YOLOv7 object detector can be seen in \Cref{fig:patch_activation_embedding}. 
The different colors in the plots correspond to the mean average precision the detector achieves 
after the bounding boxes in the image are attacked with a patch. The bright yellow dot is the 
activation of the original image without any patches present. The marker symbol of each data point 
corresponds to the parameter set used to optimize the patch. In addition to the five different 
prime patch parameter sets, a total number of 100 random noise patches and 100 grayscale patches 
are also shown.

The first experiment is about the t-distributed stochastic neighbor embeddings (t-SNE) of the
prime patches. This set of qualitative experiments provides a more general overview of the impact of
the patches on the pretrained object detector.

The embeddings of the patch activations for a single image in \Cref{fig:patch_activation_embedding}
show five distinct clusters among the different patches. The original activation of the unpatched
image is given by the yellow dot at about $(x=30, y=-13)$. Right next to this dot, the activations
of the different grayscale levels form a c-shape. The colors of the data points and the position in
the neighborhood of the original activations indicate that the impact of the grayscale patches is
low ($mAP=0.79\pm0.02$).

The second cluster with a relatively high mAP is given by the random patches. The center of this
cluster is above the first cluster at around $(20, 15)$. In contrast to the grayscale patches, the
random patches have a slightly deeper shade, which indicates that the random noise has a slightly
stronger negative impact on the detector than the grayscale patches $(0.75\pm0.01)$. At the edge is
a single prime patch of the parameter set C. 

The remaining data points of parameter set C are mixed with patches of parameter sets A and D in the
upper left cluster, with the center at around $(-12, 25)$. The shade of the cluster is darker than
the shade of the previous two clusters ($0.69\pm0.06$).

Both remaining clusters contain data points of the parameter set B. The cluster at $(-35, 5)$ is a
pure cluster of data points from parameter set B and is much denser compared to the remaining one.
Both clusters have similar mAP of $0.63\pm0.08$ and $0.63\pm0.05$, yet the last remaining cluster
with the center at around $(-10,-10)$ contains the data points with the lowest mAP values.

These observations demonstrate that noise and grayscale patches have almost no impact on the object
detector on this specific input image. Furthermore, patches that share an optimization parameter set
alter the backbone activations similarly.

\begin{figure}[tb]
\centering
\includegraphics[width=\textwidth]{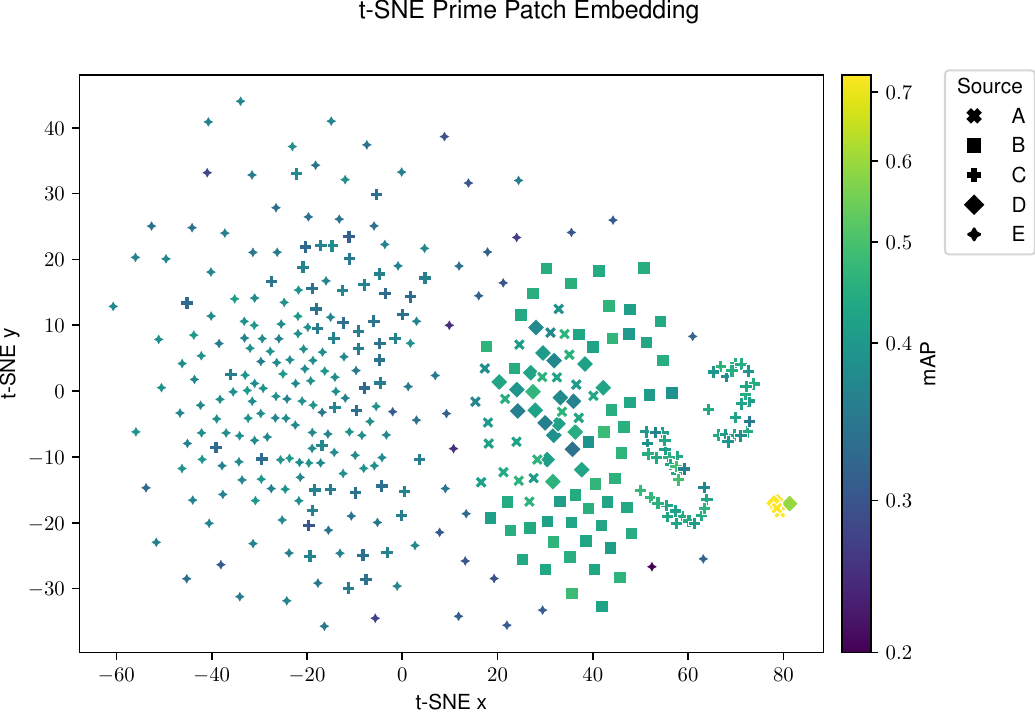}
\caption{t-SNE plot of the pixel values of each prime patch. The color indicates the impact of the
patch on the mean average precision measured with the \emph{INRIA Person} test set. The parameter
set of a patch is given by its marker symbol.}\label{fig:patch_tsne_embeddings}
\end{figure}

As this form of representation only shows the impact of a patch on a single image, another
representation is given in \Cref{fig:patch_tsne_embeddings}. Here, each data point of the t-SNE plot 
corresponds to the patch itself. The colors of the data points in this plot encode the overall 
mean average precision drop in detection performance for the \emph{INRIA Person} test set. Again,
the marker symbol corresponds to the parameter set used to optimize the prime patch.

%% file: src/exp2.tex
\subsection{Attack Performance}
\begin{figure}[tb]
\centering
\includegraphics[width=\textwidth]{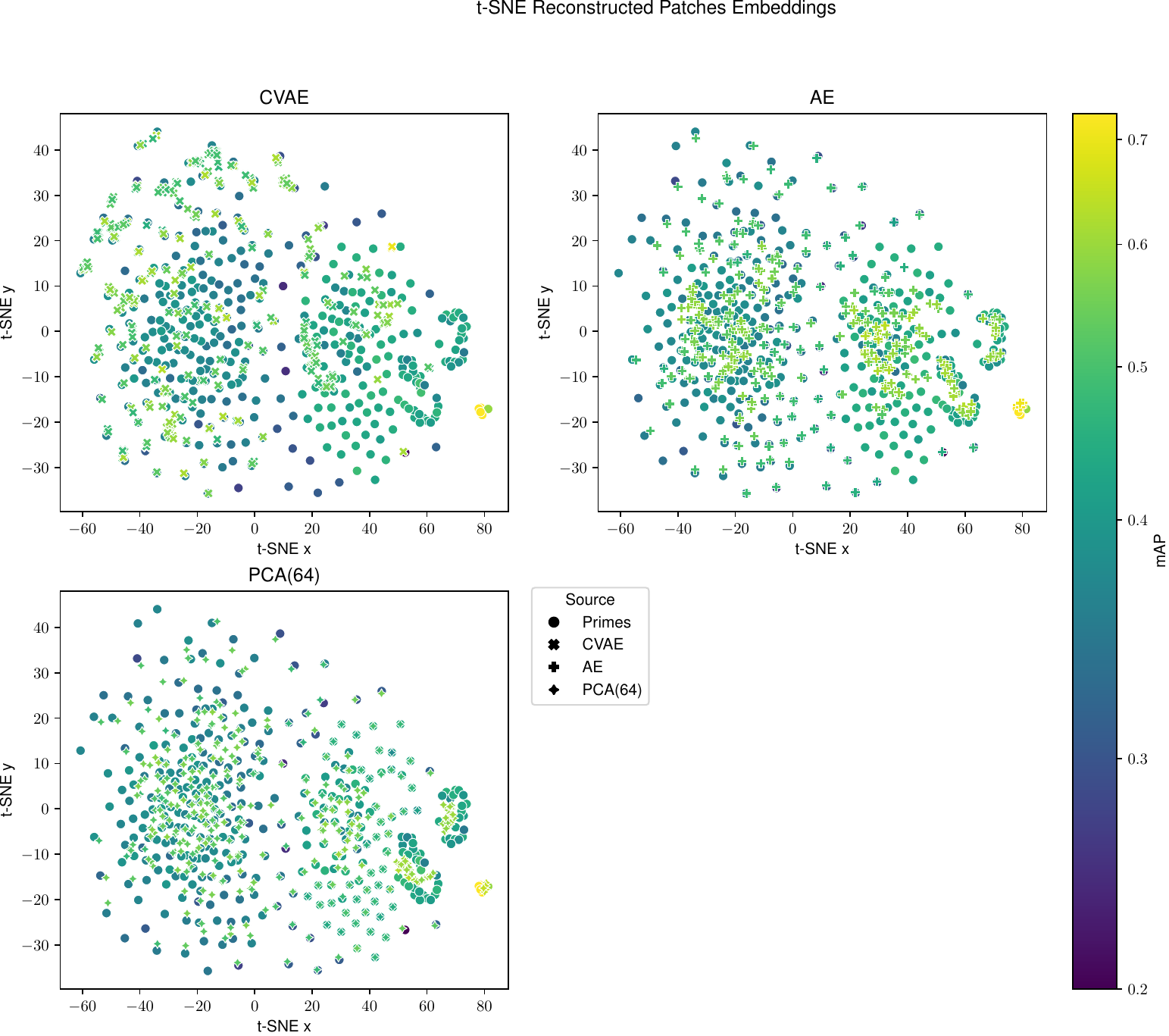}
\caption{t-SNE plots of the three reconstruction methods. Conditional Variational Autoencoder
(CVAE), Autoencoder (AE) and Principal Component Analysis with 64 components (PCA(64)). Again, the 
color indicates the impact on the mean average precision measured with the \emph{INRIA Person} test
set.  The marker symbol shows, whether it is a reconstruction from the corresponding construction
method or a prime patch.}\label{fig:patch_recreations}
\end{figure}

The next experiment covers the attack performance on the pretrained object detector when
reconstructed patches are used. In \Cref{fig:patch_recreations}, the data points correspond to
reconstructions of prime patches and prime patches themselves. Again, the color encodes the mAP
value of the corresponding patch. The marker symbol, on the other hand, provides information about
the used dimensionality reduction method.

CVAE is the conditional variational autoencoder. As the positions of the markers indicate, the 
method fails the reconstruction of the patches in most cases. Instead, multiple patches are alike.
The mean Euclidean distance of the embedded CVAE patches to its corresponding prime embedding is 
$36.81 \pm 29.31$ and the highest among all three investigated methods.

The embeddings of the auto encoder reconstructions (AE) are more scattered and less focused on a few
spots. The mean Euclidean distance of $5.27 \pm 8.58$ supports this.

The method with the lowest mean Euclidean distance of $2.96 \pm 7.10$ is the PCA with 64 components.
As shown in the plot, most prime patches can be sufficiently reconstructed. Only the patches of
parameter set E seem more challenging.

Regarding the performance of the reconstructed patches in an attack setting, all three methods are
able to reconstruct patches that result in a mean decline of the detector performance by more than
0.2 (see \Cref{tbl:reconstruction}). The prime patches and grayscale patches in the table can be
considered an upper and lower bound for the patch performance.

\begin{table*}[tb]
\centering
\caption{Mean average precision of the COCO-pretrained YOLOv7 tiny model, on the \emph{INRIA Person}
test set. The patch mode indicates, which patches are used during the attack: None is the default
without any patches. Grayscale are different grayscale levels.}\label{tbl:reconstruction}
\begin{tabular}{r l l l}
    \toprule
    Patch Mode & n & mAP 0.5 $\downarrow$ & mAP 0.5:0.95$\downarrow$ \\
    \hline
    None & 1 & $0.96$ & $0.90$ \\
    Grayscale & 11 & $0.84 \pm 0.02$ & $0.71 \pm 0.01$ \\
    Prime Patches & 375 & $0.57\pm0.06$ & $0.39\pm0.06$ \\
    \hline
    PCA (64) & 375 & $\mathbf{0.70\pm0.04}$ & $0.54\pm0.04$ \\
    AE & 375 & $0.73\pm0.04$ & $0.55\pm0.05$ \\
    CVAE& 375 & $0.72\pm0.04$ & $\mathbf{0.53\pm0.05}$ \\
    \hline
\end{tabular}
\end{table*}

%% file: src/exp3.tex
\subsection{Adversarial Training}
\Cref{tbl:adv_training} and \Cref{tbl:adv_training_crowdhuman} show the results of the adversarial
training.  Each row represents a trained network. The patch mode indicates, which method was used to
provide the adversarial patches during training. \emph{None} is the trained network without any
adversarial patches present during training.

In \Cref{tbl:adv_training}, the problem with training on a small data basis as the selected subset
of the \emph{INRIA Person} dataset becomes visible. The induced adversarial patches during training
can be considered as a form of data augmentation, resulting in a higher overall mAP. Interestingly, 
the networks are prone to grayscale patches, while at the same time, have learned to ignore or
rather profit from adversarial patches on bounding boxes of interest. A possible fix could be to
include grayscale patches during training.

The highest mAP when attacked with prime patches achieves the PCA (128) network. The best
performance without any patches visible has the PCA (16) network. The PCA (64) network has the
highest mAP when grayscale patches are visible.

When trained on the Crowdhuman dataset, the results are as expected (see
\Cref{tbl:adv_training_crowdhuman}): The induced adversarial patches during training affect the
network performance when no patches are present, and further improve the performance when patches
are present. The latter finding supports the thesis that adversarial-trained networks profit from
the patches by using them as a guidance. An adjustment of the probability that a bounding box
contains a patch during adversarial training to only $\pi=0.05$ instead of $\pi=0.25$ improves the
unpatched performance while maintaining the performance on the prime patches (see
\Cref{tbl:adv_training_crowdhuman_005}).

In comparison to the results of \Cref{tbl:adv_training}, the corresponding mAP differences between
the networks differ. The best performance, when no patches or grayscale patches are present, is
given by the unpatched network. When attacked with prime patches, both, the multipatched network and
the autoencoder network surpass the remaining networks. The PCA networks (32, 64, 128) are on par
with the single-patched network, when attacked with prime patches, yet, they perform worse when no
patches or grayscale patches are present. The lowest overall performance is given by the PCA (16)
network, followed by the CVAE network.

Due to computational intensity, the worth of training an auto encoder is questionable. Especially,
since a similar performance can be achieved with a single patch.

\begin{table*}[tb]
\centering
\caption{Mean average precision for the testset of the \emph{INRIA Person} dataset. The model first
was trained on the train data and later attacked with the given set of patches. While the YOLOv7
model has not seen any patches during the training, the other models had a 25 percent chance, that a
bounding box contained a patch during training. The patch was either a single predefined patch, a
random patch from the patch set or a sampled patch with the PCA or an autoencoder method.
}\label{tbl:adv_training}
\resizebox{\textwidth}{!}{
\begin{tabular}{r c c l l l l}
    \toprule
    & \multicolumn{2}{c}{No Patches (n=1)} & \multicolumn{2}{c}{Grayscaled (n=11)} & \multicolumn{2}{c}{Prime Patches (n=375)} \\
    Patch Mode & mAP 0.5 $\uparrow$ & mAP 0.5:0.95 $\uparrow$ & mAP 0.5 $\uparrow$ & mAP 0.5:0.95
    $\uparrow$ & mAP 0.5 $\uparrow$ & mAP 0.5:0.95 $\uparrow$ \\
    \hline
    None & $0.80$ & $0.66$ & $0.61\pm0.09$ & $0.46\pm0.12$ & $0.71\pm0.04$ & $0.57\pm0.04$ \\
    \hline
    Single-patched & $0.85$ & $0.71$& $0.71\pm0.06$ & $0.57\pm0.08$ & $0.89\pm0.03$ & $0.75\pm0.02$ \\
    Multi-patched & $0.84$ & $0.70$ & $0.71\pm0.07$ & $0.57\pm0.08$ & $0.89\pm0.02$ & $0.76\pm0.02$ \\
    PCA (16) & $\mathbf{0.85}$ & $\mathbf{0.72}$ & $0.69\pm0.07$ & $0.56\pm0.08$ & $0.86\pm0.02$ & $0.72\pm0.03$ \\
    PCA (32) & $0.82$ & $0.69$ & $0.68\pm0.07$ & $0.53\pm0.08$ & $0.85\pm0.02$ & $0.71\pm0.02$ \\
    PCA (64) & $0.85$ & $0.70$ & $\mathbf{0.73\pm0.06}$ & $\mathbf{0.59\pm0.06}$ & $0.88\pm0.02$ & $0.75\pm0.02$ \\
    PCA (128) & $0.85$ & $0.70$ & $0.72\pm0.06$ & $0.59\pm0.07$ & $\mathbf{0.90\pm0.02}$ & $\mathbf{0.76\pm0.02}$ \\
    AE & $0.84$ & $0.70$ & $0.72\pm0.06$ & $0.59\pm0.07$ & $0.87\pm0.02$ & $0.75\pm0.02$ \\
    CVAE& $0.84$ & $0.70$ & $0.70\pm0.08$ & $0.56\pm0.10$ & $0.87\pm0.02$ & $0.75\pm0.02$ \\
    \hline
\end{tabular}
}
\end{table*}

\begin{table*}[tb]
\centering
\caption{Mean average precision for the testset of the \emph{INRIA Person} dataset. The model first
was trained on the crowdhuman train data and later attacked with the given set of patches.
}\label{tbl:adv_training_crowdhuman}
\resizebox{\textwidth}{!}{
\begin{tabular}{r c c l l l l}
    \toprule
    & \multicolumn{2}{c}{No Patches (n=1)} & \multicolumn{2}{c}{Grayscaled (n=11)} & \multicolumn{2}{c}{Prime Patches (n=375)} \\
    Patch Mode & mAP 0.5 $\uparrow$ & mAP 0.5:0.95 $\uparrow$ & mAP 0.5 $\uparrow$ & mAP 0.5:0.95
    $\uparrow$ & mAP 0.5 $\uparrow$ & mAP 0.5:0.95 $\uparrow$ \\
    \hline
    None & $\mathbf{0.95}$ & $\mathbf{0.91}$ & $\mathbf{0.85\pm0.01}$ & $\mathbf{0.78\pm0.02}$ & $0.84\pm0.03$ & $0.75\pm0.05$ \\
    \hline
    Single-patched & $0.89$ & $0.79$& $0.76\pm0.04$ & $0.66\pm0.04$ & $0.93\pm0.04$ & $0.84\pm0.05$ \\
    Multi-patched & $0.84$ & $0.74$ & $0.72\pm0.02$ & $0.62\pm0.02$ & $0.95\pm0.04$ & $\mathbf{0.88\pm0.05}$ \\
    PCA (16) & $0.87$ & $0.78$ & $0.66\pm0.05$ & $0.56\pm0.05$ & $0.90\pm0.04$ & $0.80\pm0.05$ \\
    PCA (32) & $0.88$ & $0.78$ & $0.71\pm0.04$ & $0.60\pm0.04$ & $0.94\pm0.05$ & $0.85\pm0.06$ \\
    PCA (64) & $0.83$ & $0.75$ & $0.69\pm0.03$ & $0.60\pm0.02$ & $0.93\pm0.08$ & $0.86\pm0.09$ \\
    PCA (128) & $0.84$ & $0.76$ & $0.70\pm0.03$ & $0.61\pm0.03$ & $0.93\pm0.07$ & $0.86\pm0.08$ \\
    AE & $0.88$ & $0.76$ & $0.77\pm0.05$ & $0.65\pm0.06$ & $\mathbf{0.95\pm0.02}$ & $0.84\pm0.02$ \\
    CVAE& $0.88$ & $0.77$ & $0.66\pm0.06$ & $0.53\pm0.07$ & $0.92\pm0.05$ & $0.84\pm0.06$ \\
    \hline
\end{tabular}
}
\end{table*}

%% file: src/conclusion.tex
\section{Conclusion}\label{sec:conclusion}
This paper provides an in-depth analysis of adversarial patches, used to fool object detectors. To
be more specific, a set of so-called prime patches to evade a YOLOv7 based person detector is
analyzed. A qualitative insight into the activations of the backbone network is given, when the
detector is attacked with these adversarial patches.  The prime patches themselves, are processed by
three dimensionality reduction methods, and the mAP drop of their reconstructions is measured.
Furthermore, the resulting manifolds of the dimensionality reduction methods are sampled and used in
adversarial training.
The results indicate that the training of more sophisticated manifold learning methods does not
provide a significant better or more varying way to sample adversarial patches. The inclusion of a
small set of prime patches or sampled patches using a PCA is sufficient. Moreover, relying on a
diverse set of sampled patches using a learned representation results only in a small improvement
compared to naive adversarial training.

The results also show that the three investigated dimensionality reduction methods are able to
capture some of the necessary features that are required to fool an object detector. This is a
further indication that the manifold assumption applies to adversarial patches. Future work should
therefore check other manifold learning methods and use the resulting knowledge to enhance
protection mechanisms against this kind of attack.

%% file: src/appendix.tex
\begin{table}[!htb]
    \centering
    \begin{tabular}{ccccc}
    \toprule
         ID & Epochs & LR-Scheduler & Resize range & Rotation \\
         \hline
        A & 125 & StepLR & [0.5, 0.75] & 45 \\
        B & 100 & StepLR & [0.75, 1.0] & 45 \\
        C & 100 & CosineAnnealingLR & [0.75, 1.0] & 30 \\
        D & 125 & StepLR & [0.5, 0.75] & 30 \\
        E & 100 & StepLR & [0.75, 1.0] & 30 \\
    \hline
    \end{tabular}
    \caption{Different parameterization for the patch generation. The resize range is the range of 
	the scaling factor, relatively to the bounding box \cite{Bayer2023}.}
    \label{tbl:parameters}
\end{table}

\begin{figure}[!htb]
\centering
\includegraphics[width=0.19\textwidth]{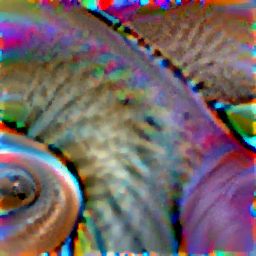}
\includegraphics[width=0.19\textwidth]{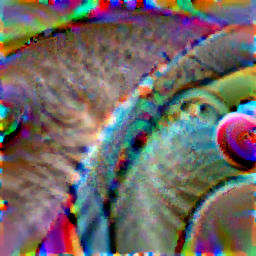}
\includegraphics[width=0.19\textwidth]{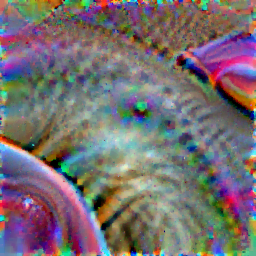}
\includegraphics[width=0.19\textwidth]{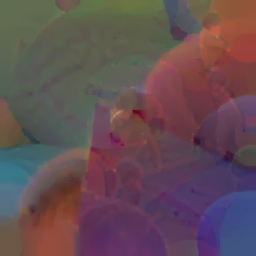}
\includegraphics[width=0.19\textwidth]{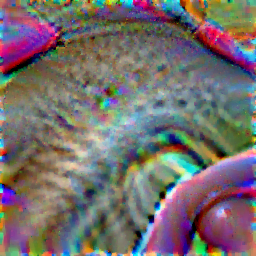}
\includegraphics[width=0.19\textwidth]{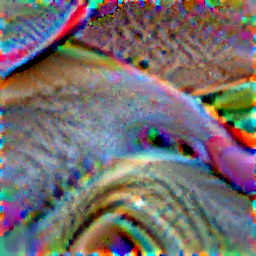}
\includegraphics[width=0.19\textwidth]{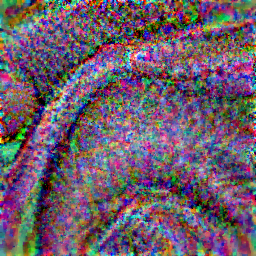}
\includegraphics[width=0.19\textwidth]{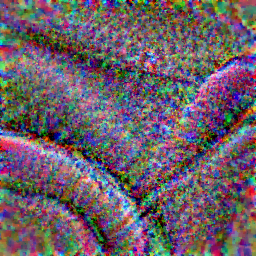}
\includegraphics[width=0.19\textwidth]{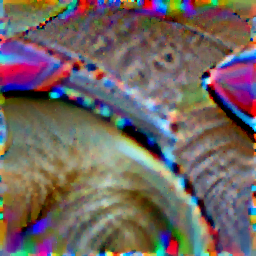}
\includegraphics[width=0.19\textwidth]{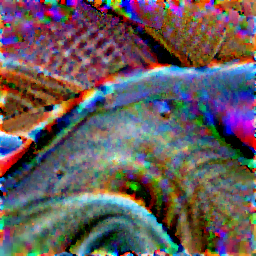}
\caption{The 10 randomly predefined prime patches used in the training for the multi-patched
network.}\label{fig:predefined_prime_patches}
\end{figure}

\begin{table}[!htb]
\centering
\caption{Mean average precision for the testset of the \emph{INRIA Person} dataset. The model first
was trained on the crowdhuman train data with $\pi=0.05$ and later attacked with the given set of patches.
}\label{tbl:adv_training_crowdhuman_005}
\resizebox{\textwidth}{!}{
\begin{tabular}{r c c l l l l}
    \toprule
    & \multicolumn{2}{c}{No Patches (n=1)} & \multicolumn{2}{c}{Grayscaled (n=11)} & \multicolumn{2}{c}{Prime Patches (n=375)} \\
    Patch Mode & mAP 0.5 $\uparrow$ & mAP 0.5:0.95 $\uparrow$ & mAP 0.5 $\uparrow$ & mAP 0.5:0.95
    $\uparrow$ & mAP 0.5 $\uparrow$ & mAP 0.5:0.95 $\uparrow$ \\
    \hline
    None & $\mathbf{0.94}$ & $\mathbf{0.89}$ & $0.82\pm0.01$ & $0.74\pm0.02$ & $0.83\pm0.04$ & $0.71\pm0.07$ \\
    \hline
    Single-patched & $0.92$ & $0.86$ & $0.83\pm0.01$ & $0.74\pm0.02$ & $0.92\pm0.02$ & $0.85\pm0.03$ \\
    Multi-patched & $0.94$ & $0.88$ & $0.83\pm0.01$ & $0.74\pm0.01$ & $\mathbf{0.95\pm0.02}$ & $\mathbf{0.89\pm0.02}$ \\
    PCA (16) & $0.91$ & $0.81$ & $0.77\pm0.03$ & $0.66\pm0.03$ & $0.92\pm0.03$ & $0.83\pm0.04$ \\
    PCA (32) & $0.93$ & $0.87$ & $0.82\pm0.01$ & $0.74\pm0.01$ & $0.93\pm0.03$ & $0.86\pm0.04$ \\
    PCA (64) & $0.93$ & $0.87$ & $0.82\pm0.01$ & $0.72\pm0.01$ & $0.93\pm0.04$ & $0.86\pm0.05$ \\
    PCA (128) & $0.94$ & $0.88$ & $0.80\pm0.01$ & $0.72\pm0.02$ & $0.93\pm0.04$ & $0.85\pm0.05$ \\
    AE & $0.93$ & $0.87$ & $\mathbf{0.85\pm0.01}$ & $\mathbf{0.77\pm0.01}$ & $0.95\pm0.02$ & $0.88\pm0.02$ \\
    CVAE& $0.93$ & $0.87$ & $0.81\pm0.01$ & $0.73\pm0.01$ & $0.93\pm0.03$ & $0.86\pm0.04$ \\
    \hline
\end{tabular}
}
\end{table}